%% file: example_paper.tex
\documentclass[nohyperref]{article}

\usepackage{bm}
\usepackage{microtype}
\usepackage{graphicx}
\usepackage{subfig}
\usepackage{booktabs} 
\usepackage{multirow}
\usepackage{adjustbox}
\usepackage[dvipsnames]{xcolor}
\usepackage{tikz}
\usetikzlibrary{positioning}
\usetikzlibrary{shapes.geometric}
\usetikzlibrary{shapes.multipart}
\usepackage{pgfplots}
\usepackage{csvsimple}

\usepackage{hyperref}

\usepackage[accepted]{icml2022}

\usepackage{amsmath}
\usepackage{amssymb}
\usepackage{mathtools}
\usepackage{amsthm}

\usepackage[capitalize,noabbrev]{cleveref}

\theoremstyle{plain}

\theoremstyle{definition}

\theoremstyle{remark}

\usepackage[textsize=tiny]{todonotes}

\icmltitlerunning{Online Partial Domain Adaptation of Convolutional Neural Networks}

\begin{document}

\twocolumn[
\icmltitle{Low-Cost On-device Partial Domain Adaptation (LoCO-PDA):\\ Enabling efficient CNN
            retraining on edge devices}



\icmlsetsymbol{equal}{*}

\begin{icmlauthorlist}
\icmlauthor{Aditya Rajagopal}{imp}
\icmlauthor{Christos-Savvas Bouganis}{imp}
\end{icmlauthorlist}

\icmlaffiliation{imp}{Intelligent Digital Systems Lab, Imperial College London, UK}

\icmlcorrespondingauthor{Aditya Rajagopal}{adityarajagopal0@outlook.com}

\icmlkeywords{Partial Domain Adaptation, CNN, Machine Learning, Efficient Training, Machine
                Learning, ICML}

\vskip 0.3in
]



\printAffiliationsAndNotice{\icmlEqualContribution} 

\begin{abstract}
    With the increased deployment of Convolutional Neural Networks (CNNs) on edge devices, the
    uncertainty of the observed data distribution upon deployment has led researchers to to utilise
    large and extensive datasets such as ILSVRC'12 to train CNNs. 
    Consequently, it is likely that the observed data distribution upon deployment is a subset of
    the training data distribution.
    In such cases, not adapting a network to the observed data distribution can cause performance
    degradation due to negative transfer and alleviating this is the focus of Partial Domain
    Adaptation (PDA).
    Current works targeting PDA do not focus on performing the domain adaptation on an
    edge device, adapting to a changing target distribution or reducing the cost of deploying the
    adapted network.
    This work proposes a novel PDA methodology that targets all of these directions and opens
    avenues for on-device PDA. 
    LoCO-PDA adapts a deployed network to the observed data distribution by enabling it to be retrained on an edge device.
    Across subsets of the ILSVRC12 dataset, LoCO-PDA improves classification accuracy by 3.04pp on
    average while achieving up to 15.1x reduction in retraining memory consumption and 2.07x
    improvement in inference latency on the NVIDIA Jetson TX2.
    The work is open-sourced at \emph{link removed for anonymity}.
\end{abstract}

\section{Introduction}
    \label{sec:intro}
    The rapid increase in the memory and compute capability of modern edge devices has resulted in
    the increased deployment of CNNs on the edge. 
    The deployment life cycle starts with the architecture and weights of the network being
    optimised on servers using large datasets such as ILSVRC'12 \cite{imagenet}.
    Once deployed on an edge device such as the NVIDIA Jetson TX2, networks typically remain
    unchanged and if adapted, require communication of data back to a server.
    This raises both data-privacy concerns and is not possible when there is no edge-server
    connectivity.

    As the deployed network is trained on a large and extensive dataset such as ILSVRC'12, it is
    expected that the label space of the observed data distribution after deployment (target
    domain) is a subset of the label space of the training data distribution (source domain). 
    However utilising this deployed network to classify solely target domain data triggers negative
    transfer \cite{pada_2018}. 
    Negative transfer refers to the loss in classification performance due to the fact that the
    network has learnt to classify a much larger number of classes (including source-only classes)
    than those in the target domain.
    Most works in the field of Partial Domain Adaptaion (PDA) aim to prevent negative transfer, but
    do not consider the possibility of a constantly changing target domain or the memory
    and compute footprint of the adaptation process or the resulting adapted network - all of which
    are important considerations for real-life applications.

    The goal of this work is to enable the partial domain adaptation of a deployed model to be
    performed directly on an edge device, within some memory and time budget, to improve the
    performance of the adapted model on the observed target domain.
    Additionally, the proposed methodology is sufficiently low-cost to allow for adaptation to a
    changing target domain and can obtain gains in inference time memory and latency footprint by
    adapting a network that is smaller than that originally deployed for inference.
    
    \input{figs_tex/introduction/pda_comparison.tex}
    The novel contribution of this work is LoCO-PDA, a methodology that uses Variational
    Autoencoders (VAEs) to perform low-cost PDA on edge devices.
    As demonstrated in Fig.\ref{fig:pda_tradeoff}, unlike other PDA methodologies, LoCO-PDA aims to
    optimise for the trade-off between domain adapted classification accuracy, and the adaptation
    process memory consumption and latency.
    \footnote{Times are collected on the server grade NVIDIA 2080Ti as SOTA methodologies are too
    memory intensive to be run on the Jetson TX2 edge device.}
    It achieves on average a 7x and 4x improvement in adaptation time and memory respectively while
    performing comparably to other SOTA PDA methodologies.

\section{Background and Related Works}
    \label{sec:background}
    \subsection{Partial Domain Adaptation (PDA)}
        \label{sec:bkgrnd_pda}
        As discussed in \cite{etn_2019}, the primary focus of unsupervised domain adaptation is
        on learning domain-invariant feature representations without access to target domain
        labels, so that the same feature extractor and classifier combination can be utilised 
        in both the source and target domains.
        However, these works \cite{uda_1_2010, uda_2_2012} assume that both source and target
        domains have the same label space, which is an unlikely scenario to be observed in
        reality.  
        A more likely scenario in real-world deployments is PDA where the target label space is
        assumed to subsume the source label space.    
        
        The challenge addressed by SOTA works targeting PDA such as ETN \cite{etn_2019}, IWAN
        \cite{iwan_2018}, and PADA \cite{pada_2018} is to train a feature extractor that generates an aligned activation
        distribution for classes of images that are common to both source and target domains,
        while preventing distribution alignment on source only classes.
        All three works utilise an adversarial framework that consists of a feature extractor
        and domain classifier.  
        The domain classifier aims to identify in an unsupervised manner the difference between
        the source and target domains, while the feature extractor is trained to reduce the
        discrepancy between the domains.
        The domain classifier outputs weights that identify classes as source-only or shared;
        and the three works vary in their approaches to utilising these weights in training the
        feature extractor.
        
        Different to such approaches, this work proposes a novel solution to the problem of PDA
        which does not change the source domain feature extractor but rather retrains just the
        classifier to better discriminate between the classes observed in the target domain.
        We propose to estimate the target domain label subspace using predictions of the deployed
        model. 
        As no ground-truth labels are used, this approach is still unsupervised.
        Furthermore, compared to adversarial PDA approaches, the proposed methodology has a low
        enough memory and compute requirement to be performed directly on an edge device and
        benefits from adaptability to changes in the target domain.

    \subsection{Edge Device CNN Training} 
        To the best of our knowledge, the only two works which also explore on-device training
        of CNNs are TinyTL \cite{tinytl_2020} and PersEPhonEE \cite{persephonee_2021}.
        TinyTL proposes a variation of the MBConv \cite{mobilenetv2_Sandler2018} layer called a
        Lite-Residual layer which downsamples the input activations, performs grouped and 1x1
        convolutions on them, upsamples the output and adds it to the output of the original MBConv
        layer.
        By only retraining the \emph{lite} bypass layers on the edge, the methodology reduces
        memory requirements of training as only downsampled intermediate activations need to be
        stored. 
        TinyTL is evaluated on a transfer-learning scenario where the source and target
        domains do not share the same label space and there is access to ground-truth labels of the
        target domain.
        This is outside the scope of this work and hence is not compared to.

        PersEPhonEE proposes an early-exit methodology instead of pruning to reduce the cost of
        both inference and training on an edge device.
        By placing an early exit after various residual blocks along ResNet50, the authors propose
        to improve both training and inference time by only retraining the desired early-exit
        block. 
        They demonstrate between 2x-22x improvement in training latency depending on the early-exit
        used and up to 20pp improvement in classification accuracy depending on the target domain.
        The authors provide results for various subsets of the ILSVRC'12 dataset and propose to use
        the output of the final classifier as labels to retrain earlier classifiers (unsupervised).
        Thus this work falls under the PDA framework and is compared against.

\section{Problem Description}
    \label{sec:problem_statement}
    The inputs to the system are a model $\mathcal{M}^0$ and a large training data
    distribution $\mathcal{D}$.
    Let $\mathcal{M}^0$ be the model that has been trained on $\mathcal{D}$ and prepared for
    initial deployment on an edge device.
    Upon deployment, the system observes a data distribution $\mathcal{D}'$ that is assumed to be
    able to change over time.
    Furthermore, the label space of $\mathcal{D}'$ is assumed to be a subset of that of
    $\mathcal{D}$. 
    Due to this assumption, it is expected that a model ($\mathcal{M}^p$), that has a lower
    memory and latency footprint than $\mathcal{M}^0$ can be adapted to the observed $\mathcal{D}'$
    after deployment as a smaller number of classes are being classified ($\mathcal{D}' \subset
    \mathcal{D}$).
    This results in an adapted network that has significant gains in inference memory and latency. 
    In this work we consider $\mathcal{M}^p$ to be a network that is obtained by pruning away $p$\%
    of the memory footprint of $\mathcal{M}^0$.
    \footnote{The pruning and retraining process is performed offline on $\mathcal{D}$. 
    Both Taylor First Order (TFO) \cite{taylor-fo_pruning_molchanov_2019} and OFA \cite{ofa_2020}
    methodologies are used for pruning.}
    However, LoCO-PDA can be applied even if $\mathcal{M}^0$ and $\mathcal{M}^p$ are completely
    different networks.
    
    The goal and novel contribution of this work is to create a system that enables the retraining
    of $\mathcal{M}^p$ on an edge device such that the accuracy of $\mathcal{M}^p$ on
    $\mathcal{D}'$ is at least as good as that of $\mathcal{M}^0$ on $\mathcal{D}'$.

\section{Low-cost Training}
    \label{sec:low-cost-training}
    \input{figs_tex/methodology/end_to_end}
    This section motivates the proposed training methodology by identifying methods to reduce the
    cost of retraining.
    The greyed out path in Fig.\ref{fig:retraining_strategy} is an abstraction of a CNN network
    architecture. 
    The feature extractor ($\mathcal{M}^p_{FE}$) takes as input an image and outputs activations 
    that are then passed to the classifier ($\mathcal{M}^p_{FC}$) to predict the class of the
    image.
    The cost of training the feature extractor is significantly larger than that of the
    classifier. 
    As stated in \cite{tinytl_2020}, the limiting factor of training CNNs on edge devices is the
    memory required to store intermediate activations used by backpropagation.
    Consequently, the primary cost of retraining a CNN lies in the feature extractor stage and
    bypassing this would significantly reduce the cost of retraining a network.
    
    One way to bypass the training of the feature extractor stage is to only train the classifier.
    This only requires inference, which is a highly optimised process on edge devices.
    However, doing so also requires the storage of images with which retraining can be performed.
    This can be infeasible on edge devices due to the limited available memory.
    Consequently, LoCO-PDA aims to retrain just the classifier but also bypass the execution of the
    feature extractor, thereby significantly reducing the memory requirements of the retraining
    process. 
     

\section{Methodology}
    \label{sec:methodology}
    Most modern networks have linear classifier layers as they have highly expressive feature
    extractors that create representations of images (activations) that can be linearly separated.
    LoCO-PDA exploits the simplicity of these activations and uses light-weight VAEs to generate
    them instead of performing inference through the feature extractor of $\mathcal{M}^{p}$. 
    These generated activations can then be used to train the classifier.
    This process is described by the black path in Fig.\ref{fig:retraining_strategy}.
    Utilising a pruned model achieves gains in inference memory consumption and latency, while
    retraining the classifier allows the pruned network to regain the accuracy lost by pruning
    while also optimising the network's weights to the observed data distribution. 
    There is no added overhead to performing inference through $\mathcal{M}^0$ as this is the
    originally deployed model which is being used for inference and thus statistics of
    $P(y_t|\mathcal{X}_t)$ can be collected over time before initiating retraining.
    
    \subsection{VAE Training Process}
        \label{sec:vae_train_proc}
        \input{figs_tex/methodology/vae_training}
        LoCO-PDA uses an extension of VAEs known as Conditional-VAEs (CVAEs). 
        VAEs \cite{vae_2014} are generative models that learn to model some distribution $P(t)$
        where $t$ are high-dimensional data points which can represent a variety of inputs.
        They do this by learning a latent representation that can then be sampled to generate data
        points close $P(t)$. 
        Fig.\ref{fig:vae_training} shows the process of training the CVAEs to generate the
        activations of $\mathcal{M}^p$ at the output of the feature extractor.
        CVAEs \cite{cvae_2015} allow for the sampling of the latent space to be conditioned on a
        desired class of $\mathcal{A}_s$.
        One CVAE is required per source domain dataset $\mathcal{D}$ and model $\mathcal{M}^{p}$
        and they are trained on $\mathcal{M}^{p}$'s activations for all images ($\mathcal{X}_s$) in
        $\mathcal{D}$ before deployment.
        
        
        The encoder in Fig.\ref{fig:vae_training} learns the function $q_{\phi}(z |
        \mathcal{A}_s, y_s)$ which maps activations to latent space variables.
        The decoder then learns the function $p_{\theta}(\mathcal{A}_s | z, y_s)$ which
        generates an activation corresponding to provided class.
        It learns to generate activations for all source domain classes ($\mathcal{Y}_s$) in order
        to be able to adapt to varying target domains.
        
        The loss function used is that proposed in \cite{beta_vae_2017} for conditional VAEs and is
        given by: 
        \begin{equation}
            \begin{aligned}
                \mathcal{L}(\theta, \phi; \mathcal{A}_s, z, y_s, \beta) = &
                            \mathbb{E}_{q_{\phi}(z | \mathcal{A}_s, y_s)}
                            [\textrm{log}\, p_\theta(\mathcal{A}_s | z, y_s)] \\
                &- \beta\ \mathcal{KL}(q_\phi(z | \mathcal{A}_s, y_s) || p(\bm z)) 
            \end{aligned}
        \end{equation}
        The weights of the encoder and decoder that are learnt are given by $\phi$ and $\theta$
        respectively. 
        $\beta$ is a parameter that balances the trade-off between reconstruction error and latent
        space regularisation.
        $p(\bm z)$ is the prior distribution and is set to $\mathcal{N}(\bm 0, \bm I)$ in order to
        obtain a closed form solution to the regularisation term.
        The conditional decoding process samples $\bm z \sim \mathcal{N}(\bm 0, \bm I)$, augments
        the variable with $y_s$ corresponding to the desired class and provides this as input to
        the trained decoder as shown in Fig.\ref{fig:retraining_strategy}.
        This process is facilitated by learning a latent space that closely follows
        $\mathcal{N}(\bm 0, \bm I)$ (KL-divergence loss term), thus training a decoder that can map
        a randomly sampled variable in this distribution to the desired network activation based on
        the conditioning information.\footnote{Refer to Appendix
        \ref{app:vae_training_hyperparameters} for the training hyperparameters.}
    
    \subsection{Identifying $\mathcal{D}'$}
        Once the VAE is trained, it is deployed along with the initial deployment model
        $\mathcal{M}^0$ and the pruned model $\mathcal{M}^p$ that will be adapted to the observed
        data distribution.
        Before retraining $\mathcal{M}^p$ on $\mathcal{D}'$, it is necessary to identify the labels
        observed. 
        We propose to identify these labels from the distribution predicted by the currently
        deployed network $\mathcal{M}^0$ (Fig.\ref{fig:retraining_strategy}).
        The assumption here is that $\mathcal{M}^0$ has sufficient accuracy to identify the classes
        observed upon deployment. 
    
    \subsection{Retraining $\mathcal{M}^p$}
        The retraining process gets as input a distribution $P(y_t|\mathcal{X}_t)$
        which assigns a probability per observed class based on the frequency with which that class
        was observed.
        The tunable parameter $R$ decides how many total activations to use when performing the
        classifier retraining. 
        The $R$ activations that are generated follow the class distribution
        $P(y_t|\mathcal{X}_t)$, thus a sufficiently small probability could result in no activation
        being generated for that class.
        Under the assumption that $\mathcal{M}^0$ predicts the correct class with a sufficiently
        high probability, this approach can be robust to noisy class distributions generated by
        $\mathcal{M}^0$ that may contain incorrect class predictions.  
        These activations are then used to train $\mathcal{M}^{p}$'s classifier.
        Compared to the feature extractor, the decoder can be up to 20x smaller in terms of memory
        footprint which reduces the cost of retraining such that it can be performed on a CPU
        within a reasonable time budget.
        \footnote{Refer to Appendix \ref{app:retraining_hyperparams} for the retraining
        hyperparameters}
        
\section{Evaluation}
    \label{sec:evaluation}
    This section first evaluates LoCO-PDA under the constraints of a practical deployment scenario.
    Sec.\ref{sec:eval_pda} then compares LoCO-PDA to other SOTA PDA methodologies. 
    
    Consider the following scenario, as proposed by \cite{perf4sight_2021}, of an autonomous
    vehicle running an image classification CNN on an NVIDIA Jetson TX2 edge device. 
    As the environment the car observes varies over time, the weights of a pruned version of the
    deployed CNN can be adapted to specialise this network to the data observed. 
    The Jetson TX2 has only 8GB of memory that is shared between the CPU and GPU.
    Apart from the image classification network, there are other safety critical applications
    running on the device.
    
    This scenario introduces the following considerations. 
    As the device has a unified memory system, memory intensive processes such as retraining
    started on either processor can affect processes from executing on the other.
    In safety critical scenarios, this could have catastrophic consequences.
    On-device retraining of CNNs is primarily limited by its memory consumption and
    for many CNNs even batch sizes as low as 32 requires too much memory.
    Additionally, the robustness of the methodology to the lack of ground-truth labels of the
    observed data is an important consideration.
    
    
    Consequently all approaches in this section are evaluated on the following metrics.
    Static (storage) and runtime memory consumption.
    Classification accuracy of the adapted network on $\mathcal{D}'$.
    Performance under noisy $\mathcal{D}'$ estimation.
    Wall-clock time to perform on-device adaptation on both the GPU and CPU.
    Inference latency and total floating-point operations (FLOPs) of the adapted network.

    \subsection{Evaluation Datasets and Methodologies}
        \label{sec:eval_datasets_methods}
        This section details the methodologies that will be evaluated and the datasets that they
        will be evaluated on. 
        For all experiments in this section, the training hyperparameters provided in Appendix
        \ref{app:vae_training_hyperparameters} and \ref{app:non_vae_retraining_hyperparameters}
        were used.
        
        \subsubsection{Datasets}
            \label{sec:eval_datasets}
            For all experiments in this section, the source domain dataset ($\mathcal{D}$) on which
            model training and pruning are performed is the entire ILSVRC'12 training set.
            
            For the practical deployment scenario, adaptation to four example target domain
            datasets ($\mathcal{D}'$) will be explored.
            These are constructed as subsets of the ImageNet dataset that emulate different
            environments that an autonomous vehicle could encounter \cite{perf4sight_2021}.
            They are City (185 classes), Motorway (26 classes), Country-side (204 classes),
            Off-road (26 classes).
            Furthermore, for this evaluation, an edge-retraining dataset is created by randomly
            sampling 20\% of the ImageNet training dataset.
            For each $\mathcal{D}'$, images from this dataset are utilised to emulate images that a
            deployed network would classify and are used to retrain the network on the edge. 
            Although the deployed network has already observed these images during training on
            ImageNet, a different random seed is used for image preprocessing and augmentation to
            emulate  variation from the training distribution.  
            Doing so allows for all accuracy results to be reported on the full ImageNet validation
            dataset (unseen by models) to allow for fair comparison in the future.
            
            When evaluating the methodology against other resource unconstrained PDA approaches,
            the Caltech-84 and Office-31 datasets are used. 
            The Caltech-84 dataset uses the 84 classes that are shared between the
            ILSVRC'12 and Caltech-256 \cite{caltech256} datasets.
            The Office-31 dataset consists of 3 categories corresponding to images from
            \emph{Amazon}, a \emph{DSLR} and a \emph{Webcam} and each category has images
            corresponding to 31 classes.
            These target domains are common transfer learning benchmarks and details of the classes
            present in all target domains are provided in Appendix
            \ref{app:target_domain_datasets}.
        
        \subsubsection{Methodologies}
            \label{sec:eval_methods}
            The methodologies that LoCO-PDA is compared against under the practical deployment
            scenario setting store a set of $N$ possible pruned models on-device, each of which can
            be retrained to the observed data distribution.
            The choice of model that is retrained depends on the target inference memory and
            latency at the time of deployment. 
            The methodologies vary in either pruning strategy or the manner in which multiple
            networks with different budgets are exposed.

            \textbf{On-device Retraining} A set of $N$ pruned models are obtained using both Taylor
            First Order \cite{taylor-fo_pruning_molchanov_2019} and OFA pruning \cite{ofa_2020}.
            All pruning is performed using the tool open sourced in \cite{dapr_rajagopal2020}.
            For both pruning strategies, on device retraining on $\mathcal{D}'$ of all-layers and
            classifier-only are considered as baselines.

            \textbf{PersEPhonEE} This methodology appends early-exit modules which contain two
            convolution layers and a classifier at various points along the backbone network.
            After training of all exits and the backbone network offline on $\mathcal{D}$,
            on-device retraining cost is minimised by only finetuning the early exit that falls
            within the available memory and compute budget.
            
            To the best of our knowledge, PersEPhonEE is the only other work that explores edge
            device retraining for PDA.
            The TFO and OFA approaches are baselines to evaluate the robustness of LoCO-PDA to
            different pruning strategies.
            
            For all methodologies, both memory bounded and unbounded training approaches are
            evaluated.
            This refers to placing a memory limit on the number of images that can be stored in
            order to perform retraining versus utilising the entire retraining dataset which as
            discussed in Sec.\ref{sec:eval_datasets} corresponds
            to 20\% of the ILSVRC'12 training dataset.
            In the memory bounded scenario, a limit of 400MB is set as this is deemed to be a
            reasonable allocation of the available 8GB on a Jetson TX2 for this purpose. 
            In the memory unbounded scenario, as each $\mathcal{D}'$ has a different number of
            classes, the storage memory utilised varies and is 9.40 GB, 1.36 GB, 10.3 GB and 1.36
            GB for the City, Motorway, Country-side, and Off-road $\mathcal{D}'$ respectively.
            The memory unbounded scenario evaluates the best case performance of each methodology
            when not subjected to any memory constraints.
            All retraining based approaches are evaluated on the NVIDIA Jetson TX2 using PyTorch
            v1.6, CUDA 10.2, cuDNN 8.0 and batch-size 32.
            Under the scenario where ground-truth labels in the target domain are used, these
            methodologies act as an "oracle" for retraining a network directly on the target
            domain. 

            For resource unconstrained PDA, LoCO-PDA is compared against ETN \cite{etn_2019}, IWAN
            \cite{iwan_2018} and PADA \cite{pada_2018}.

        \subsubsection{Networks}
            \label{sec:eval_networks}
            \input{tables_tex/evaluation/networks_summary.tex}
            The $\mathcal{M}^0$ and $\mathcal{M}^p$ networks utilised in this evaluation are shown
            in Table.\ref{tab:network_summary}.
            As discussed in Sec.\ref{sec:eval_methods}, $N$ $\mathcal{M}^{p}$ networks are stored
            on device, however for simplicity this section only evaluates the case of $N=1$.
            
            $\mathcal{M}^0_{TFO}$ is an unpruned ResNet50 network and $\mathcal{M}^{80}_{TFO}$ is
            the same network TFO pruned by 80\%.
            $\mathcal{M}^0_{OFA}$ is the largest sub-network that can be sampled from the OFA super
            network and $\mathcal{M}^{80}_{OFA}$ is an OFA sub-network that has a
            similar FLOPs budget to $\mathcal{M}^{80}_{TFO}$.
            $\mathcal{M}^0_{PE}$ is an unpruned ResNet50 network augmented with 6 early exits
            located at FLOPs normalised locations along the network.
            Consequently, apart from \emph{model memory}, all other metrics are the same as
            $\mathcal{M}^0_{TFO}$ in Table.\ref{tab:network_summary}.
            $\mathcal{M}^{80}_{PE}$ is the second early exit (after layer2.1) as this has a similar
            FLOPs budget to $\mathcal{M}^{80}_{TFO}$.
            \footnote{Refer to Appendix \ref{app:further_experiments} for results on other
            architectures.}

    \subsection{Static and Runtime Memory Footprint}
        \label{sec:eval_memory_evaluation}
        \input{tables_tex/evaluation/static_runtime_memory.tex}
        Static memory footprint of a methodology corresponds to the storage memory consumed by the
        various parts of the methodology required to perform network adaptation.
        Runtime memory footprint corresponds to the memory consumed during network adaptation.
        Note that this is the footprint of the network adaptation process and not that of performing
        inference on the network. 
        The total static and runtime memory consumption of the various methodologies on a Jetson
        TX2 is displayed in Table.\ref{tab:static_runtime_memory}. 
        
        \paragraph{TFO, OFA: Image based retraining} For each methodology (TFO, OFA), the static
        network storage component consists of the corresponding $\mathcal{M}^0$ to estimate labels
        and $\mathcal{M}^{80}$ which is adapted through retraining.
        Note depending on the number ($N$) of pruned networks ($\mathcal{M}^{p}$) stored, the
        static network storage component will increase.
        Here, the case of $N=1$ is evaluated.   
        The static image storage component consists of 400MB of images for memory bounded 
        retraining.
        The runtime memory consumption is affected by whether classifier-only or all-layer
        retraining is performed and is quoted for retraining $\mathcal{M}^{80}$ on the GPU of
        the Jetson TX2.
        
        \paragraph{PersEPhonEE: Early-exit retraining} The network component of the static memory
        consumption is from $\mathcal{M}^{80}_{PE}$.
        The images component of the static memory consumption is from the 400 MB of images
        stored to perform memory bounded early-exit retraining. 
        The runtime component is from performing on-device retraining of the second early exit
        (Sec.\ref{sec:eval_networks}) on the GPU. 
        
        \paragraph{LoCO-PDA} For LoCO-PDA, the additional static network component is storage of
        the CVAE model used to generate activations.
        This component is 9.65 MB for $\mathcal{M}^{80}_{TFO}$ and 16.83 MB for
        $\mathcal{M}^{80}_{OFA}$.
        As PersEPhonEE uses $\mathcal{M}^{0}_{TFO}$ as the backbone network, the CVAE trained
        on activations generated by $\mathcal{M}^{80}_{TFO}$ is used for comparison.
        No images are used for retraining with this approach, hence the static image component
        is 0 MB.
        The memory consumed by the generated activations is incorporated into the runtime
        memory consumption which is evaluated both on the CPU and GPU of the Jetson TX2. 
        The significantly extra memory consumption of the GPU is due to CUDA initialisation
        overheads which can be close to 1 GB depending on the device and framework
        combination. 
        LoCO-PDA executed on a CPU consumes 13.3x, 8.76x and 15.1x less memory compared to TFO
        classifier-only retraining, OFA classifier-only retraining and second exit training of
        PersEPhonEE respectively executed on a GPU. 

    \subsection{Accuracy of $\mathcal{M}^p$ on $\mathcal{D}'$}
        \label{sec:eval_accuracy}
        \input{tables_tex/evaluation/unlim_memory_consumption.tex}
        This section compares the best achieved Top1-Test accuracy after adapting the network to
        the four autonomous driving $\mathcal{D}'$s for all the methodologies discussed in
        Sec.\ref{sec:eval_methods}.
        In order to evaluate the methodologies under a no-uncertainty situation, all results here
        assume knowledge of the ground-truth labels for the images in $\mathcal{D}'$.
        Later sections evaluate the effect of uncertain labels when using $\mathcal{M}^0$ to
        identify $\mathcal{D}'$. 
        First, memory bounded and unbounded; and all-layers and classifier-only training approaches
        are evaluated. 
        \subsubsection{Memory bounded vs. unbounded}
            \label{sec:eval_mem_lim_vs_unlim}
            As seen in Table \ref{tab:train_approach_comparison}, the memory unbounded approach outperforms the memory bounded approach by 2.64pp on
            average across $\mathcal{D}'$ and methods.
            Given the small performance difference (2.64pp) and the large memory requirements (3.4x) of the memory unbounded approach vs. the memory bounded approach, the rest of the evaluation focuses on the memory bounded case.
        
        \subsubsection{All layers vs. Classifier-only}
            Retraining all layers performs on average across $\mathcal{D}'$ 0.2pp worse than
            classifier-only training. 
            Furthermore, retraining all layers has a runtime memory footprint that is 1.8x and
            training latency that is 3.4x larger than classifier-only training.
            Hence the following sections use classifier-only retraining as a baseline.

        \subsubsection{LoCO-PDA}
            \input{tables_tex/evaluation/per_subset_acc_comparison.tex}
            The \emph{No Label Space Uncertainty} columns of Table.\ref{tab:acc_comparison} display
            the achieved Test Top1 accuracy on each autonomous driving $\mathcal{D}'$ for
            $\mathcal{M}^{80}_{TFO}$ and $\mathcal{M}^{80}_{OFA}$ when 
            \emph{no retraining}, \emph{memory bounded classifier-only retraining} and
            \emph{LoCO-PDA} are performed on-device.
            PersEPhonEE was not compared against here due to its poor performance
            (Table.\ref{tab:train_approach_comparison}). 
            This poor performance is due to utilising predictions from an early exit placed only
            after the second Bottleneck block of ResNet50 to allow for a FLOPs normalised
            comparison between methodologies (\ref{sec:eval_networks}).

            The following results are averages across both the OFA and TFO networks. 
            LoCO-PDA consistently outperforms memory bounded retraining of $\mathcal{M}^{80}$ by
            1.46pp on average across all $\mathcal{D}'$.
            Furthermore, on the smaller $\mathcal{D}'$ (Motorway, Off-road), on which domain
            adaptation is more beneficial, LoCO-PDA also outperforms its corresponding unpruned
            network ($\mathcal{M}^0$) by 3.04pp on average.
            Intuitively, larger subsets do not benefit significantly from on-device adaptation as
            the initial network optimised on many classes learns a classifier that best separates a
            space with many classes.
    
    \subsection{Storage Memory Limitations}

\input{figs_tex/evaluation/storage_mem_lim.tex}
        The results from the previous sections utilise a storage memory limit of 400MB for image
        based retraining approaches. 
        However, as discussed in Sec.\ref{sec:eval_mem_lim_vs_unlim} placing this restriction did
        have a reduction in accuracy when compared to the memory unbounded scenario. 
        This section expands on this to demonstrate the effect of various storage limits on the
        achieved accuracy when performing image based classifier-only retraining.
        The red line in Fig.\ref{fig:storage_mem} displays the achieved classifier-only retraining
        accuracy for $\mathcal{M}^{80}_{TFO}$ on the Off-road $\mathcal{D}'$ as the available
        memory to store images is varied (x-axis).
        These values are the average of five retraining runs.
        The solid green line displays the no retraining accuracy of $\mathcal{M}^{80}_{TFO}$ and the
        solid blue line displays the accuracy achieved by LoCO-PDA.
        The dashed blue line is the static memory consumed by LoCO-PDA.
        The graphs show that as the available memory budget is increased, the achieved retraining
        accuracy improves and depending on $\mathcal{D}'$ can outperform LoCO-PDA under no label
        uncertainty while utilising 6.8x times the memory.
        For the same memory budget as LoCO-PDA, the memory-bounded retraining approach
        performs on average 4.88pp worse than LoCO-PDA.

    \subsection{Noisy $\mathcal{D}'$ Estimation}
        \label{sec:eval_noisy_labels}
        All results presented thus far assume knowledge of the label space of
        $\mathcal{D}'$ which is not representative of real deployment scenarios.
        In this work we propose to estimate the label space of $\mathcal{D}'$ using the predictions
        of the initially deployed model $\mathcal{M}^{0}$.
        As the accuracy of this model is not 100\%, some of the labels obtained from this model are
        incorrect.
        This section evaluates the effect on achieved adaptation accuracy of noisy $\mathcal{D}'$
        estimation. 
        The following results are averages across both the OFA and TFO networks. 
        
        The \emph{Label Space Uncertainty} columns in Table.\ref{tab:acc_comparison} show the
        achieved test top1 accuracy on $\mathcal{D}'$ for both the memory bounded classifier-only
        retraining strategy (\textsc{Retrained}) and the LoCO-PDA approach.
        For the image based retraining approaches, memory bounded retraining with uncertain labels
        causes a 7.27pp degradation in achieved accuracy compared to retraining with certain labels
        across $\mathcal{D}'$.
        Furthermore in most cases, retraining with noisy labels actually degrades the network's
        accuracy compared to $\mathcal{M}^{80}$ without any retraining.
        On the other hand, LoCO-PDA only incurs a 0.67pp reduction in achieved accuracy across
        $\mathcal{D}'$ and never a degradation compared to $\mathcal{M}^{80}$ without retraining. 

        With image based retraining, predicting an incorrect label and utilising this combination
        to retrain the network causes increased confusion as the network is now being trained to
        associate incorrect labels and images.    
        As LoCO-PDA generates the activations corresponding to the estimated class, only an
        incorrect ratio of the labels is generated and the retraining stage still associates labels
        with their correct corresponding activations.
        Thus, LoCO-PDA is significantly more robust to noisy estimations of $\mathcal{D}'$ than
        image based retraining approaches.

    \subsection{On-device Adaptation Wall-clock Time}
        \label{sec:eval_adaptation_time}
        \input{tables_tex/evaluation/wall-clock.tex}
        This section evaluates the wall-clock time to perform on-device adaptation using each
        methodology on the Jetson TX2.
        The values in Table.\ref{tab:wall_clock_time} are the time taken to retrain the
        pruned network until the accuracy under the \emph{No Label Space Uncertainty} case in
        Table.\ref{tab:acc_comparison} is achieved.
        As different $\mathcal{D}'$ behave differently under retraining, the quoted values are
        averaged across the four $\mathcal{D}'$.  
        The time taken to perform on-device adaptation for image based retraining approaches is
        infeasible on a CPU, hence the GPU times are utilised here for comparison.
        The results show that LoCO-PDA executed on a CPU and GPU outperform classifier-only
        retraining on a GPU by 2.9x and 12.9x respectively.
        
        By bringing domain adaptation latencies down to less than a minute, while consuming less
        than 300 MB of memory and being significantly robust to noisy estimations in
        $\mathcal{D}'$, LoCO-PDA opens up avenues for on-device partial domain adaptation of
        networks that have not been explored before.

    \subsection{Partial Domain Adaptation}
        \label{sec:eval_pda}
        \input{tables_tex/evaluation/pda.tex}
        This section compares LoCO-PDA against SOTA PDA methodologies ETN \cite{etn_2019}, IWAN
        \cite{iwan_2018}, and PADA \cite{pada_2018}.
        As these works do not consider the case of pruning a network, the reported results in this
        section are for an unpruned ResNet50. 
        
        For the ILSVRC'12 to Caltech-84 transfer setting, activations corresponding to the 84
        classes shared between the ILSVRC'12 and Caltech-256 datasets are generated using a CVAE
        trained on ILSVRC'12 data that has no knowledge of the target Caltech-84 activation
        distribution.
        Furthermore, the target distribution is estimated using the output of classifying the
        entire Caltech-84 dataset on the unpruned ResNet50 network, hence no ground-truth labels
        were used.
        For the Office-31 setting, ResNet50 is first finetuned to each of the three categories
        using all 31 classes and 3 CVAEs are trained to generate activations for each of these
        source domain networks.
        The target domain images are then classified on the source domain finetuned network to
        obtain a distribution and the corresponding CVAE is used to generate activations to retrain
        the network.

        On average across all PDA settings, Table.\ref{tab:pda} shows that LoCO-PDA improves
        performance compared to no domain adaptation and PADA by 5.42pp and
        1.20pp respectively; and performs comparably to IWAN (only 0.95pp worse).
        Although performing 3.43pp worse compared to ETN, LoCO-PDA has the significant advantage
        over all three methodologies that it can rapidly adapt to changes in the target
        domain. 
        PADA, IWAN and ETN cannot be performed on an edge device to re-adapt a network to new
        domains due to their large memory and latency footprints (Fig.\ref{fig:pda_tradeoff}).

\section{Conclusion}
    \label{sec:conclusion}
    This work proposes LoCO-PDA, a methodology to perform low cost on-device partial domain
    adaptation of CNNs along with an open-sourced implementation.
    In the case where a pruned version of a network is adapted to the observed target domain,
    the proposed methodology achieves up to 3.27x reduction in model memory footprint and 2.07x
    improvement in inference latency while achieving on average 3.04pp improvement in
    classification accuracy compared to an unpruned version of the network that is deployed without
    any domain adaptation.
    Compared to other image based on-device retraining methodologies, LoCO-PDA achieves up to 15.1x
    runtime memory consumption improvements while also being significantly more robust to noisy
    estimations of the target domain class
    distribution.
    The entire retraining process can be performed on a CPU with sub-minute domain adaptation times
    on a Jetson TX2 device, opening avenues for on-device PDA which have not been explored before.
    LoCO-PDA also performs comparably to other SOTA PDA methodologies while benefiting from the
    ability to adapt to varying target domains, which the other methodologies cannot.

\bibliographystyle{apalike}
\bibliography{example_paper}
\newpage
\appendix
\input{appendix.tex}

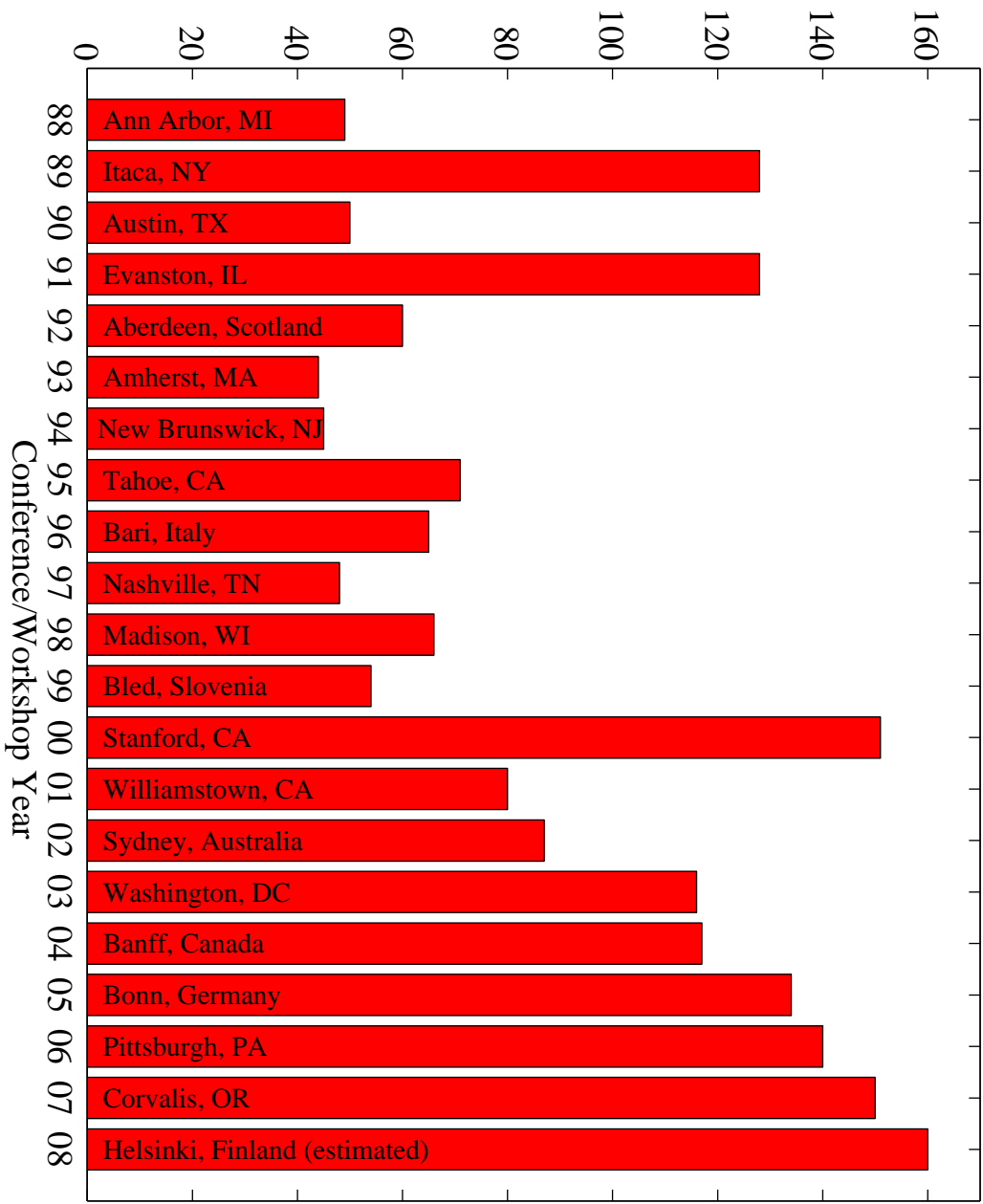
\end{document}

%% file: figs_tex/introduction/pda_comparison.tex
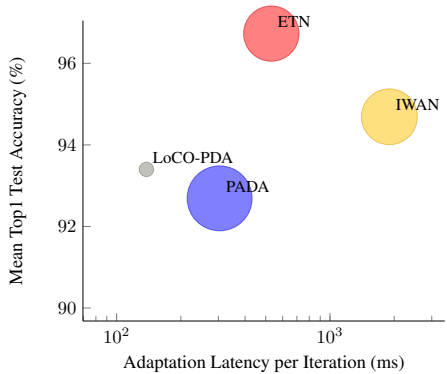
\begin{figure}[t]
\centering
\begin{minipage}[b]{0.35\textwidth}
\resizebox{\textwidth}{!}{
    \begin{tikzpicture}
        \begin{axis}[
            xmode = log,
            xlabel = Adaptation Latency per Iteration (ms),
            ylabel = Mean Top1 Test Accuracy (\%),
            xtick pos = left,
            ytick pos = left,
            enlarge x limits = {abs=7pt},
            enlarge y limits = {abs=7pt},
            axis x line*=bottom,
            axis y line*=left,
            ymin = 90,
            xmin = 80,
            xmax = 3000
            ]
            
            \addplot [scatter, 
                      only marks,
                      mark=*,
                      opacity=0.5,
                      visualization depends on = {value \thisrow{method} \as \methodname},
                      nodes near coords*={\methodname},
                      nodes near coords align={anchor=south west},
                      every node near coord/.append style={black, font={\small}, opacity=1.0},
                      visualization depends on = {value \thisrow{memory}/500 \as \markersize},
                      scatter/@pre marker code/.append style={/tikz/mark size = \markersize},
                      scatter/@post marker code/.append style={},
                      ] table [
                    y=acc,
                    x=latency, 
                    col sep=comma
                ] {figs_tex/introduction/pda_comparison_per_iter.csv};
        \end{axis}
    \end{tikzpicture}
}
\end{minipage}
\caption{Adaptation process latency per iteration (x-axis), memory (dot-area) and mean accuracy
            (y-axis) trade-off across PDA transfer tasks on the Office31 dataset for SOTA PDA
            methodologies and LoCO-PDA.
            Please note the logarithmic x-axis.}
\label{fig:pda_tradeoff}
\vspace{-0.5cm}
\end{figure}

%% file: figs_tex/methodology/end_to_end.tex
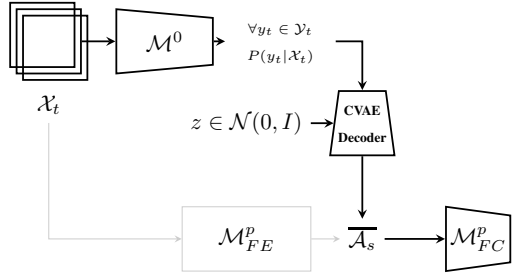
\begin{figure}[t]
\centering
\begin{minipage}[b]{0.40\textwidth}
\resizebox{\textwidth}{!}{
\begin{tikzpicture}[node distance = 1.8cm,thick,font=\bfseries]
    \node[draw,
        minimum width = 1cm,
        minimum height = 1cm
        ] () at (0,0) {};
    \node[draw,
        minimum width = 1cm,
        minimum height = 1cm
        ] (images) at (0.1,-0.1) {};
    \node[draw,
        minimum width = 1cm,
        minimum height = 1cm,
        ] () at (0.2,-0.2) {};
    \node[below = 0.2cm of images] (X_label) {$\mathcal{X}_t$}; 
        
    \node[trapezium, 
        draw, 
        right of = images,
        rotate = -90,
        minimum width = 1cm,
        minimum height = 1.5cm,
        trapezium stretches body,
        ] (m0) {};
    \node[] at (m0.center) {$\mathcal{M}^0$};

    \draw[-stealth] (images.east) -- (m0.south);

    \node[right of = m0] (m0_preds) {
        \begin{tabular}{c}
            \tiny{$\forall y_t \in \mathcal{Y}_t$} \\
            \tiny{$P(y_t | \mathcal{X}_t)$}
        \end{tabular}};
    \draw[-stealth] (m0.north) -- (m0_preds.west);

    \node[trapezium,
        draw,
        below right of = m0_preds,
        minimum width = 1cm,
        minimum height = 1cm,
        trapezium stretches body,
        ] (vae_decoder) {};
    \node[] at (vae_decoder.center){
        \begin{tabular}{c} 
            \tiny{CVAE}\\ 
            \tiny{Decoder} 
        \end{tabular}};

    \draw[-stealth] (m0_preds.east) -| (vae_decoder.north);

    \node[left of = vae_decoder] (z) {$z \in \mathcal{N}(0,I)$};
    \draw[-stealth] (z.east) -- (vae_decoder.west);

    \node[below of = vae_decoder] (acts) {$\overline{\mathcal{A}_s}$};
    \draw[-stealth] (vae_decoder.south) -- (acts.north);

    \node[trapezium,
        draw,
        right of = acts,
        rotate = -90,
        minimum width = 1cm,
        minimum height = 1cm,
        trapezium stretches body
        ] (student_fc) {};
    \node[] at (student_fc.center) {$\mathcal{M}^p_{FC}$};

    \draw[-stealth] (acts.east) -- (student_fc.south);

    \begin{scope}[thin]
        \node[draw=gray!40,
            minimum width = 2cm,
            minimum height = 1cm,
            left of = acts,
            ] (student_fe) {$\mathcal{M}^p_{FE}$};
        
        \draw[-stealth, gray!40] (X_label.south) |- (student_fe.west);
        \draw[-stealth, gray!40] (student_fe.east) -- (acts.west);
    \end{scope}

\end{tikzpicture}
}
\end{minipage}
\caption{
    Proposed methdology for low-cost adaptation of the pruned network's classifier
    ($\mathcal{M}^p_{FC}$).
    $\mathcal{M}^0$ produces a set of predicted labels ($\mathcal{Y}_t$), for target domain images
    $\mathcal{X}_t$.
    The predicted classes along with their frequency of occurrence ($P(y_t | \mathcal{X}_t)$) is  provided to the trained decoder.
    The decoder then generates a set of estimated activations
    $\overline{\mathcal{A}_s}$ which can be used to train the pruned network's classifier. 
    Consequently, this process bypasses the feature extractor of the pruned network ($\mathcal{M}^p_{FE}$).
} 
\label{fig:retraining_strategy}
\vspace{-0.5cm}
\end{figure}

%% file: figs_tex/methodology/vae_training.tex
\begin{figure}[t]
\centering
\begin{minipage}[b]{0.4\textwidth}
\resizebox{\textwidth}{!}{
\begin{tikzpicture}[node distance = 1.8cm,thick,font=\bfseries]
    \node[draw,
        minimum width = 1cm,
        minimum height = 1cm
        ] () at (0,0) {};
    \node[draw,
        minimum width = 1cm,
        minimum height = 1cm
        ] (images) at (0.1,-0.1) {};
    \node[draw,
        minimum width = 1cm,
        minimum height = 1cm,
        ] () at (0.2,-0.2) {};
    \node[below = 0.2cm of images] (X_label) {$\mathcal{X}_s$}; 

    \node[draw,
        minimum width = 2cm,
        minimum height = 1cm,
        right of = images,
        ] (student_fe) {$\mathcal{M}^p_{FE}$};
    
    \draw[-stealth] (images.east) -- (student_fe.west);
        
    \node[right of = student_fe] (acts) {$\mathcal{A}_s$};
    \draw[-stealth] (student_fe.east) -- (acts.west);
    
    \node[trapezium,
        draw,
        below right = 0.8 and 1.4 of acts,
        rotate = -180,
        minimum width = 1cm,
        minimum height = 1cm,
        trapezium stretches body,
        ] (vae_encoder) {};
    \node[] at (vae_encoder.center){
        \begin{tabular}{c} 
            \tiny{CVAE}\\ 
            \tiny{Encoder} 
        \end{tabular}};

    \draw[-stealth] (acts.east) -- ++(0.3,0) -| (vae_encoder.south);
    
    
    \node[left = 0.4 of vae_encoder.20] (src_labels) {\small{$y_s \in \mathcal{Y}_s$}};
    \draw[-stealth] (src_labels.east) -- (vae_encoder.20);
    
    \node[below = 0.25 of vae_encoder.110] (var) {$\Sigma$}; 
    \node[below = 0.3 of vae_encoder.65] (mean) {$\mu$}; 
    \draw[-stealth] (vae_encoder.65) -- (mean.north);
    \draw[-stealth] (vae_encoder.110) -- (var.north);

    \node[circle,
        draw,
        scale = 0.5,
        below = 0.2 of var] (mul) {$\times$};
    \node[circle,
        draw,
        scale = 0.5,
        left = 0.2 of mul] (add) {$+$};

    \node[right = 0.2 of mul] (n) {$n \in$ \small{${\mathcal{N}(0,\mathcal{I})}$}};
    \node[left = 0.2 of add] (z) {$z$};

    \draw[-stealth] (n.west) -- (mul.east);
    \draw[-stealth] (mul.west) -- (add.east);
    \draw[-stealth] (mean.south) -- (add.north);
    \draw[-stealth] (var.south) -- (mul.north);

    \node[trapezium,
        draw,
        below of = student_fe,
        rotate = -90,
        minimum width = 1cm,
        minimum height = 1cm,
        trapezium stretches body,
        ] (vae_decoder) {};
    \node[] at (vae_decoder.center){
        \begin{tabular}{c} 
            \tiny{CVAE}\\ 
            \tiny{Decoder} 
        \end{tabular}};

    \draw[-stealth] (add.west) -- (z.east); 
    \draw[-stealth] (z.west) -- ++(-0.3,0) |- (vae_decoder.70);
    \draw[-stealth] (src_labels.south) |- (vae_decoder.110);

    \node[left of = vae_decoder] (recon_x) {$\overline{\mathcal{A}_s}$};
    \draw[-stealth] (vae_decoder.south) -- (recon_x.east);

    \begin{scope}[thin, blue!40]
        \node[rectangle,
            draw,
            below = 0.9 of vae_decoder] (recon_loss) 
                                        {$|| \overline{\mathcal{A}_s} - \mathcal{A}_s ||^2_2$};
        \node[rectangle,
            draw,
            right = 0.9 of recon_loss] (kld_loss) 
                        {$\mathcal{KL}[\mathcal{N}(\mu,\Sigma)||\mathcal{N}(0,\mathcal{I})]$};
    \end{scope}

\end{tikzpicture}
}
\end{minipage}
\caption{
    VAE training process used to train the decoder that is used in
    Fig.\ref{fig:retraining_strategy}.
    The whole process if performed offline using source domain images ($\mathcal{X}_s$) and
    available ground truth source domain labels ($\mathcal{Y}_s$).
    The CVAEs are trained to reproduce source domain activations ($\mathcal{A}_s$) when provided
    with the desired class $y_s$ and a random vector $z$ sampled from a multivariate standard
    normal distribution.
    The blue boxes show the loss functions used to train the CVAEs, where the left box is the
    reconstruction mean-squared-error (MSE) between the estimated activations
    $\overline{\mathcal{A}_s}$ and the true activations $\mathcal{A}_s$ and the right box is the
    KL-divergence between the learnt distribution of the latent space ($\mathcal{N}(\mu,\Sigma)$)
    and $\mathcal{N}(0,\mathcal{I})$.
}
\label{fig:vae_training}
\vspace{-0.4cm}
\end{figure}
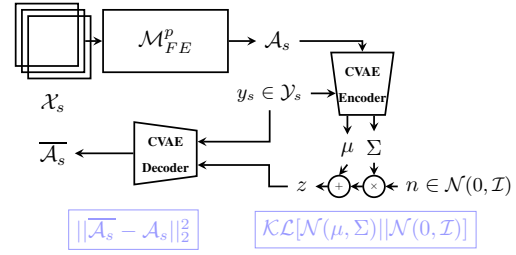

%% file: tables_tex/evaluation/networks_summary.tex
\begin{table}[t]
    \begin{minipage}[b]{\linewidth}
    \caption{Summary of inference operations (\textsc{Ops}), Model memory footprint (\textsc{Model}), Jetson TX2 single image inference latency (\textsc{Lat}) and ILSVRC'12 Test Top1 accuracy (\textsc{Acc})}
    \begin{small}
    \resizebox{\textwidth}{!}{
        \begin{tabular}{llcccc}
            \toprule
            & & \multirow{3}{*}{\textbf{\textsc{Ops}}} & \multirow{3}{*}{\textbf{\textsc{Model}}} & \multirow{3}{*}{\textbf{\textsc{Lat}}} & \multirow{3}{*}{\textbf{\textsc{Acc}}}\\
            & & & & & \\
            & & \textbf{\textsc{(MFLOPs)}}             & \textbf{\textsc{(MB)}}                   & \textbf{(ms)}                     & \textbf{\textsc{(\%)}} \\
            \midrule
            \multirow{2}{*}{\textbf{\textsc{TFO}}}          & $\mathcal{M}^{0}_{TFO}$  & 4087 & 102 & 45.13 & 76.12 \\
                                                            & $\mathcal{M}^{80}_{TFO}$ & 1250 & 41  & 36.66 & 63.74 \\
            \midrule
            \multirow{2}{*}{\textbf{\textsc{OFA}}}          & $\mathcal{M}^{0}_{OFA}$  & 3496 & 126 & 42.34 & 83.89 \\
                                                            & $\mathcal{M}^{80}_{OFA}$ & 1197 & 40  & 20.44 & 81.75 \\
            \midrule
            \multirow{2}{*}{\textbf{\textsc{PerseEPhonEE}}} & $\mathcal{M}^{0}_{PE}$   & 4087 & 143 & 45.13 & 76.12 \\
                                                            & $\mathcal{M}^{80}_{PE}$  & 1246 & 143 & 16.65 & 16.91 \\
            \bottomrule
        \end{tabular}
    }
    \end{small}
    \label{tab:network_summary}
    \end{minipage}
    \vspace{-0.5cm}
\end{table}

%% file: tables_tex/evaluation/static_runtime_memory.tex
\begin{table}[t]
    \begin{minipage}[b]{\linewidth}
    \caption{Static and runtime memory consumption of the various methodologies.
    \textsc{Network} refers to the model storage requirements. 
    \textsc{Images} refers to the storage of images for retraining.}
    \begin{small}
    \resizebox{\textwidth}{!}{
        \begin{tabular}{llcccr}
            \toprule
            & & \multicolumn{2}{c}{\textbf{\textsc{Static (MB)}}} & \multirow{2}{*}{\textsc{\textbf{Runtime}}} & \multirow{2}{*}{\textsc{\textbf{Total}}} \\
            \cmidrule(lr){3-4}
            & & \textsc{Network} & \textsc{Images} & \textbf{\textsc{(MB)}} & \textbf{\textsc{(MB)}} \\
            \midrule
            \multirow{3}{*}{\textbf{\textsc{TFO}}} & Classifier-only & \multirow{2}{*}{143} & \multirow{2}{*}{400} & 1668 & 2211 \\
                                                   & All-Layers      &                      &                      & 3508 & 4052 \\
                                                   & \textbf{LoCO-PDA (CPU)} & \multirow{2}{*}{153} & \multirow{2}{*}{0}  & 13   & \textbf{166} \\
                                                   & LoCO-PDA (GPU)          &                      &                     & 1143 & 1296            \\
            \midrule
            \multirow{3}{*}{\textbf{\textsc{OFA}}} & Classifier-only & \multirow{2}{*}{166} & \multirow{2}{*}{400}       & 1865 & 2431 \\
                                                   & All-Layers      &                      &                            & 2800 & 3366 \\
                                                   & \textbf{LoCO-PDA (CPU)}  & \multirow{2}{*}{249}      & \multirow{2}{*}{0}    & 28      & \textbf{277} \\
                                                   & LoCO-PDA (GPU)           &                           &                       & 1128    & 1337            \\
            \midrule
            \multicolumn{2}{c}{\textbf{\textsc{PersEPhonEE}}} & 143 & 400 & 1971 & 2513 \\
                                                    
            \bottomrule
        \end{tabular}
    }
    \end{small}
    \label{tab:static_runtime_memory}
    \end{minipage}
\end{table}

%% file: tables_tex/evaluation/unlim_memory_consumption.tex

\begin{table}[t]
    \begin{minipage}[b]{\linewidth}
    \caption{Best achieved Top1-Test accuracy for the various baseline approaches to on-device retraining. Values are averaged over all four autonomous driving $\mathcal{D}'$.}
    \begin{small}
    \resizebox{\textwidth}{!}{
        \begin{tabular}{lccccc}
            \toprule
            & \multicolumn{2}{c}{\textbf{\textsc{All Layers}}} & \multicolumn{3}{c}{\textbf{\textsc{Classifier Only}}} \\
            \cmidrule(lr){2-3}
            \cmidrule(lr){4-6}
            & \textsc{$\mathcal{M}^{80}_{TFO}$} & \textsc{$\mathcal{M}^{80}_{OFA}$} & \textsc{$\mathcal{M}^{80}_{TFO}$} & \textsc{$\mathcal{M}^{80}_{OFA}$} & \textsc{$\mathcal{M}^{80}_{PE}$} \\
            \midrule
            \textbf{\textsc{Memory Bounded}}   & 76.88 & 81.44 & 75.15 & 82.83 & 23.79 \\
            \textbf{\textsc{Memory Unbounded}} & 81.19 & 83.14 & 79.22 & 84.63 & 26.70 \\
            \bottomrule
        \end{tabular}
    }
    \end{small}
    \label{tab:train_approach_comparison}
    \end{minipage}
\end{table}

%% file: tables_tex/evaluation/per_subset_acc_comparison.tex
\begin{table*}[t]
    \begin{minipage}[b]{\linewidth}
    \caption{On-device domain adaptation accuracy of $\mathcal{M}^{80}_{TFO}$ and $\mathcal{M}^{80}_{OFA}$. 
            \textsc{Retrained} is the 400MB memory bounded, classifier-only retraining strategy. 
            The number of classes in each $\mathcal{D}'$ is provided in brackets.
            Results show label space certain and uncertain scenarios.}
    \begin{small}
    \resizebox{\textwidth}{!}{
        \begin{tabular}{lcc ccc ccc cc cc}
            \toprule
            & & & \multicolumn{6}{c}{\textsc{No Label Space Uncertainty}} & \multicolumn{4}{c}{\textsc{Label Space Uncertainty}} \\ 
            \cmidrule(lr){4-9}
            \cmidrule(lr){10-13}
            & \textsc{$\mathcal{M}^{0}_{TFO}$ (\%)} & \textsc{$\mathcal{M}^{0}_{OFA}$ (\%)} & \multicolumn{3}{c}{\textsc{$\mathcal{M}^{80}_{TFO}$ (\%)}} & \multicolumn{3}{c}{\textsc{$\mathcal{M}^{80}_{OFA}$ (\%)}} & \multicolumn{2}{c}{\textsc{$\mathcal{M}^{80}_{TFO}$ (\%)}} & \multicolumn{2}{c}{$\mathcal{M}^{80}_{OFA}$ (\%)} \\
            \cmidrule(lr){4-6}
            \cmidrule(lr){7-9}
            \cmidrule(lr){10-11}
            \cmidrule(lr){12-13}
            & \textsc{No Retrain} & \textsc{No Retrain} & \textsc{No Retrain} & \textsc{Retrained} & \textbf{\textsc{LoCO-PDA}} & \textsc{No Retrain} & \textsc{Retrained} & \textbf{\textsc{LoCO-PDA}} & \textsc{Retrained} & \textbf{\textsc{LoCO-PDA}} & \textsc{Retrained} & \textbf{\textsc{LoCO-PDA}}\\
            \midrule
            \textbf{\textsc{City} (185)}           & 78.01 & 81.64 & 70.89 & 71.11 & 73.84 & 78.50 & 80.37 & 80.73 & 66.36 & 72.33 & 76.59 & 80.67 \\
            \textbf{\textsc{Motorway} (26)}        & 74.54 & 78.28 & 68.23 & 71.28 & 76.75 & 74.85 & 79.91 & 79.86 & 53.93 & 75.20 & 73.49 & 79.05 \\
            \textbf{\textsc{Country-side} (204)}   & 78.48 & 81.98 & 71.36 & 71.78 & 74.03 & 79.07 & 80.28 & 81.22 & 67.82 & 73.04 & 78.12 & 80.56 \\
            \textbf{\textsc{Off-road} (26)}        & 82.87 & 85.99 & 76.20 & 86.43 & 87.30 & 82.32 & 90.78 & 89.94 & 72.13 & 86.88 & 85.38 & 90.55 \\
            \bottomrule
        \end{tabular}
    }
    \end{small}
    \label{tab:acc_comparison}
    \end{minipage}
\end{table*}

%% file: figs_tex/evaluation/storage_mem_lim.tex
\begin{figure}[t]
\centering
\begin{minipage}[b]{0.3\textwidth}
\resizebox{\textwidth}{!}{
    \begin{tikzpicture}
        \begin{axis}[
            xmode = log,
            xlabel = Static Memory Used (MB),
            ylabel = Top1 Test Accuracy (\%),
            xtick pos = left,
            ytick pos = left,
            axis y line*=left,
            axis x line*=bottom,
            enlarge x limits = {abs=7pt},
            enlarge y limits = {abs=7pt},
            legend cell align = {left},
            legend style = {draw=none, at={(0.005,0.87)}, anchor=north west}
            ]
            
            \addplot [
                mark=none, 
                color=blue,
                ] coordinates {(1,87.30)(10000,87.30)};
            \addlegendentry{LoCO-PDA}
            
            \addplot table [
                x=mb_used, 
                y=top1, 
                mark=none, 
                col sep=comma
                ] {figs_tex/evaluation/off_road.csv};
            \addlegendentry{$\mathcal{M}_{TFO}^{80}$ Retrained}
    
            \addplot[
                mark=none,
                dotted, thick,
                color=blue,
                ] coordinates {(166,75.0)(166,88.0)};
            \addlegendentry{LoCO-PDA Mem}
    
            \addplot[
                mark=none,
                color=black!30!green
                ] coordinates {(1,76.20)(10000,76.20)};
            \addlegendentry{No Retrain $\mathcal{M}_{TFO}^{80}$}
            
            \addplot[
                mark=none,
                dotted, thick,
                color=black!10!orange
                ] coordinates {(1125,75.0)(1125,88.0)};
            
            \node (src) at (axis cs:166,82.0) {};
            \node (dst) at (axis cs:1125,82.0) {};
            \draw [<->] (src) -- node[below]{6.8x} (dst);
        \end{axis}
    \end{tikzpicture}
}
\end{minipage}
\caption{Effect of available image storage memory on achieved retraining accuracy for the Off-road
            $\mathcal{D}'$.
            Please note the logarithmic scale on the x-axis.}
\label{fig:storage_mem}
\vspace{-0.5cm}
\end{figure}
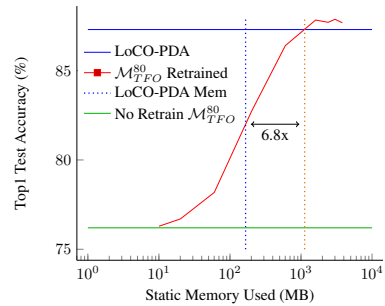

%% file: tables_tex/evaluation/wall-clock.tex
\begin{table}[t]
    \begin{minipage}[b]{\linewidth}
    \caption{The on-device network adaptation time for various methodologies averaged across $\mathcal{D}'$.
                The quoted time in min:sec is the time taken to achieve the accuracy provided in
                Table.\ref{tab:acc_comparison}.}
    \begin{small}
    \resizebox{\textwidth}{!}{
        \begin{tabular}{l cc cc}
            \toprule
            & \multicolumn{2}{c}{\textbf{\textsc{TFO}}} & \multicolumn{2}{c}{\textbf{\textsc{OFA}}} \\
            \cmidrule{2-3}
            \cmidrule{4-5}
            & \textsc{Retrained} & \textbf{\textsc{LoCO-PDA}} & \textsc{Retrained} & \textbf{\textsc{LoCO-PDA}} \\
            \midrule
            \textbf{\textsc{GPU (min:sec)}} & 02:43  & 00:12 & 02:02  & 00:10 \\
            \textbf{\textsc{CPU (min:sec)}} & 329:53 & 00:46 & 344:00 & 00:54 \\
            \bottomrule
        \end{tabular}
    }
    \end{small}
    \label{tab:wall_clock_time}
    \vspace{-0.4cm}
    \end{minipage}
\end{table}

%% file: tables_tex/evaluation/pda.tex
\begin{table}[t]
    \begin{minipage}[b]{\linewidth}
    \caption{Comparison of LoCO-PDA to SOTA PDA approaches.
                The domains are Caltech-84 (C84), Amazon (A), DSLR (D) and Webcam (W).
                The ResNet50 column corresponds to a ResNet50 network trained on the source domain
                without further adaptation.}
    \begin{small}
    \resizebox{\textwidth}{!}{
        \begin{tabular}{ll c ccc c}
            \toprule
            & & \multirow{2}{*}{\textbf{\textsc{ResNet50}}} & \multicolumn{3}{c}{\textbf{\textsc{SOTA}}} & \multirow{2}{*}{\textbf{\textsc{LoCO-PDA}}} \\
            \cmidrule{4-6}
            & & & ETN & IWAN & PADA & \\
            \midrule
            \multicolumn{2}{l}{ILSVRC'12 $\rightarrow$ C84} & 69.69 & 83.23 & 78.06 & 75.03 & 79.17 \\
            \midrule
            \multirow{6}{*}{\textsc{Office31}} & A $\rightarrow$ W & 76.03 & 94.52 & 89.15 & 86.54 & 84.64 \\
                                               & A $\rightarrow$ D   & 91.12 & 95.03 & 90.45 & 82.17 & 90.56 \\
            \cmidrule{2-7}
                                               & D $\rightarrow$ W   & 94.06 & 100.0 & 99.32 & 99.32 & 97.50 \\
                                               & D $\rightarrow$ A   & 83.62 & 96.21 & 95.62 & 92.69 & 93.74 \\
            \cmidrule{2-7}
                                               & W $\rightarrow$ A & 88.42 & 94.64 & 94.26 & 95.41 & 93.95 \\
                                               & W $\rightarrow$ D  & 98.69 & 100.0 & 99.36 & 100.0 & 100.0 \\
            \bottomrule
        \end{tabular}
    }
    \end{small}
    \label{tab:pda}
    \vspace{-0.5cm}
    \end{minipage}
\end{table}

%% file: appendix.tex
\section{Hyperparameters}
    \subsection{VAE Training Hyperparameters}
        \label{app:vae_training_hyperparameters}
        As described in \cite{beta_annealing_2015}, the value of $\beta$ changes over the course of
        training and is initialised at 0 (favouring reconstruction only) and gradually increased in
        steps of $\delta_{\beta}$ every $ep_{\beta}$ epochs of training.
        Furthermore, the training process has the hyperparameters of $bs^{vae}$ (training batch
        size), $lr_0^{vae}$ (inital learning rate), $\gamma^{vae}$ (learning rate multiplier),
        $ep^{vae}_{lr}$ (learning rate step size). 
        
        
        \subsubsection{Training Hyperparameters}
            For all experiments in the following sections unless mentioned otherwise, the following
            values are used for the introduced hyperparameters: 
            \begin{itemize}
                \item $|z| = 16$ (size of the latent space), $s=1000$ (number of classes in ILSVRC'12)
                \item $q_\phi(z|\mathcal{A}_s,y_s)$ is estimated using a four layer fully
                    connected neural network with layer sizes [$|\mathcal{A}_s| + s$, 1024],
                    [1024, 128], [128, 64], [64, $|z|$] with ReLU non-linearities.
                \item $p_\theta(\mathcal{A}_s|z,y_s)$ is estimated using a two layer fully
                    connected neural network with layer sizes [$|z| + s$, 512], [512,
                    $|\mathcal{A}_s|$] with ReLU non-linearities. 
                \item Adam \cite{adam_2015} optimiser with $bs^{vae} = 512$, $lr_0^{vae} =
                    1e^{-3}$, $\gamma^{vae} = 0.1$, $ep^{vae}_{lr} = 30$ - This is set such that
                    the learning rate only changes after the value of $\beta$ has reached 1.
                \item 90 epochs of training are performed in total.
            \end{itemize}
           
    \subsection{Retraining Hyperparameters}
        \label{app:retraining_hyperparams}
        \begin{itemize}
            \item $R = 3000$
            \item Stochastic Gradient Descent (SGD) optimiser with 50 epochs of retraining,
                batch size 32, initial learning rate of $1e^{-6}$, learning rate changes every
                15 epochs with $\gamma = 0.1$, momentum of 0.9 and no weight decay.
        \end{itemize}
    
    \subsection{On-device Retraining Hyperparameters}
        \label{app:non_vae_retraining_hyperparameters}
        For the TFO and OFA methodologies, 10 epochs of retraining are performed with an initial
        learning rate of $1e^{-3}$ which is decayed by 0.1 every 3 epochs. 
        For PersEPhonEE, as discussed in \cite{persephonee_2021}, 10 epochs of retraining are
        performed with an initial learning rate of $1e{-2}$ that is left unchanged.

\section{Further Experiments}
    \label{app:further_experiments}
    This section discusses additional experiments performed to identify the limits of LoCO-PDA.
    
    \subsection{Other network architectures}
        \label{sec:fe_other_network_architectures}
        \input{tables_tex/further_experiments/mobilenet_v2_50_acc.tex}
        \subsubsection{MobileNetV2 \cite{mobilenetv2_Sandler2018}}
            The results from Sec.\ref{sec:evaluation} demonstrate that LoCO-PDA generalises well
            across different pruning strategies (TFO, OFA) for the same backbone network
            (ResNet50).
            This section evaluates LoCO-PDA's generalisability to other networks by evaluating it
            on MobileNetV2.
            Table.\ref{tab:mobv2_acc_comparison} displays the accuracy regained by performing
            on-device memory bounded, classifier-only training and LoCO-PDA on the four autonomous
            driving $\mathcal{D}'$. 
            $\mathcal{M}^0_{TFO}$ is an unpruned MobileNetV2 with 312.3 MFLOPs of inference cost
            and 14.09 MB of model memory.
            $\mathcal{M}^{50}_{TFO}$ is a 50\% TFO pruned MobileNetV2 network with 166.6 MFLOPs of
            inference cost and 7.06 MB of model memory.
            
            Across all $\mathcal{D}'$, LoCO-PDA performs 0.88pp better than memory bounded
            classifier-only retraining.
            However, although the image based retraining methodology degrades accuracy by 9.01pp in
            the label uncertain setting compared to label certain setting, LoCO-PDA also degrades
            accuracy by 2.45pp. 
            This could be due to the lower accuracy of the unpruned MobileNetV2 compared to the
            networks in Sec.\ref{sec:evaluation}, thus demonstrating the sensitivity of LoCO-PDA to
            an overly noisy estimation of $\mathcal{D}'$.

        \subsubsection{VGG11 \cite{vgg_Simonyan_15}}
            All the networks evaluated thus far and all modern CNN architectures have linear
            classifiers while AlexNet \cite{alexnet} and VGG11 \cite{vgg_Simonyan_15} have
            non-linear classifiers. 
            Consequently, utilising a CVAE, with significantly lower expressive power than a
            feature extractor, to estimate the activations of networks such as AlexNet and VGG is
            challenging. 
            Furthermore, the dimensionality of the input to VGG11's feature extractor has 10,388
            neurons making it challenging to create a CVAE architecture with a reasonable memory
            footprint to estimate so many neurons.  
            Preliminary results fail to successfully train small VAEs to estimate the activations
            of VGG11 and further exploration of this limitation is left for future work.
    
    \subsection{Unconditional vs Conditional VAEs}
        \label{sec:fe_uncon_con}
        \input{tables_tex/further_experiments/uncon_con.tex}
        Instead of training a single CVAE for all 1000 classes of ILSVRC'12, it is possible to
        train smaller unconditional VAEs per class and depending on the activation desired, the
        appropriate VAE is sampled.  
        The architecture of the VAE used is provided in Appendix \ref{app:vae_archs}, and each VAE
        consumes only 0.4 MB. 
        This experiment explores if conditioning the VAE on classes causes any loss in
        representation capabilities.
        Table.\ref{tab:uncon_con_comparison} shows the results for no label space uncertainty
        retraining of $\mathcal{M}^{80}_{TFO}$ for various $\mathcal{D}'$ using LoCO-PDA.
        The results demonstrate that on average across $\mathcal{D}'$, the unconditional approach
        performs 0.22pp worse than one CVAE while consuming 40x more memory.
        Thus, a conditional architecture has significant memory benefits and similar representation
        capabilities as the unconditional architecture. 

\section{Architectures, Hyperparameters}
    \label{app:vae_archs}
    For all experiments other than those in Appendix \ref{sec:fe_uncon_con}, the VAE architecture
    described in Appendix \ref{app:vae_training_hyperparameters} is used.
    For the experiment in Appendix \ref{sec:fe_uncon_con}, the CVAE follows this same architecture
    however the architecture of the VAE is as follows:
    \begin{itemize}
        \item $m = 2$, $s=1000$ (number of classes in ILSVRC'12)
        \item $\mathcal{A}_{q}$ is a three layer fully connected neural network with layer sizes
            [$n + s$, 128], [128, 64], [64, $m$] with ReLU non-linearities.
        \item $\mathcal{A}_{p}$ is a two layer fully connected neural network with layer sizes [$m
            + s$, 64], [64, $n$] with ReLU non-linearities.
    \end{itemize}
    
    For all experiments other than those in Sec.\ref{sec:eval_pda}, the LoCO-PDA hyperparameters
    provided in Appendix \ref{app:vae_training_hyperparameters} were used.
    For the results provided in Sec.\ref{sec:eval_pda}, training schedule followed as as follows:
    \begin{itemize}
        \item SGD optimiser with 5 epochs of retraining, batch size 32, initial learning rate
            $1e^{-6}$, with no learning rate changes, momentum of 0.9 and no weight decay.
    \end{itemize}

\section{Target Domain Datasets ($\mathcal{D}'$)}
    \label{app:target_domain_datasets}
    This section provides details on the classes present in all the source and target domains
    discussed in Sec.\ref{sec:evaluation}. 
    The details of the classes present in the four autonomous driving $\mathcal{D}'$ are:
    \begin{itemize}
        \item \textbf{City} (185 classes) : ambulance, trash can, station wagon, tandem bike, taxi,
            car mirror, car wheel, convertible, 
                crane, electric locomotive, fire engine, fire truck, garbage truck, petrol pump,
                radiator grille, jeep landrover, rickshaw, lawn mower, limousine, mailbox, manhole
                cover, minibus, minivan, Model T, moped, scooter, mountain bike, moving van,
                parking meter, passenger car / coach, 
                pay-phone, pickup truck, police car, race car, school bus, shopping cart, snowplow,
                sports car, steel arch bridge, tram, suspension bridge, tow truck, trolley bus,
                street sign, traffic light, palace, 
                mosque, church building, castle, boathouse, tirumphal arch, academic gown/robe,
                cardigan, fur coat, gown, jersey / t-shirt, suit, sunglasses, sweatshirt, trench
                coat, umbrella, swan, dogs, red fox, cats

        \item \textbf{Motorway} (26 classes) : station wagon, bullet train, car mirror, car wheel,
            convertible, electric locomotive, petrol 
                pump, jeep landrover, minibus, minivan, mobile home, Model T, moving van, 
                passenger car / coach, pay-phone, pickup truck, police car, race car, recreational
                vehicle, snowplow, sports car, tow truck, trailer truck, street sign, water tower

        \item \textbf{Country-side} (204 classes) : wheelbarrow, station wagon, bullet train, taxi,
            car mirror, car wheel, convertible, 
                electric locomotive, freight car, garbage truck, petrol pump, radiator grille, 
                half track, horse cart, jeep landrover, lawn mower, mailbox, manhole cover,
                minibus, minivan, mobile home, Model T, moped, scooter, mountain bike, moving van,
                oxcart, parking meter, passenger car / coach, 
                pay-phone, picket-fence, pickup truck, plough, police car, race car, recreational
                vehicle, school bus, snowmobile, snowplow, sports car, steel arch bridge, tank,
                thatched roof, tile roof, tow truck, tractor, 
                trailer truck, worm fence, street sign, traffic light, hay, palace, mosque, church
                building, castle, lighthouse, barn, viaduct, water tower, cardigan, fur coat, gown,
                sarong, jersey / t-shirt, suit, sunglasses, 
                sweatshirt, swimming trunks, trench coat, umbrella, cock, hen, quail, goose, swan,
                dogs, red fox, cats, rabbits, ram, sheep

        \item \textbf{Off-road} (26 classes) : mobile home, mountain bike, oxcart, pickup truck,
            plough, snowmobile, tank, tractor, hay, 
                ostrich, iguana, alligator, wallaby, koala, wombat, brown bear, black bear, hog, 
                wild boar, ox, water buffalo, bison, wild deer 
    \end{itemize}
    There are fewer written categories than the number of classes stated as some categories such as
    "dogs" have many ILSVRC'12 classes within them.
    There is also overlap of classes between the subsets as would be expected.
    For the Office-31 dataset, the 10 classes (which are the same for each of the 3 categories of
    images) that made up the various $\mathcal{D}'$ are:
        \begin{itemize}
            \item back\_pack, bike, calculator, headphones, keyboard, laptop\_computer, monitor,
                mouse, mug, projector
        \end{itemize}

%% file: tables_tex/further_experiments/mobilenet_v2_50_acc.tex
\begin{table}[t]
    \begin{minipage}[b]{\linewidth}
    \begin{small}
    \resizebox{\textwidth}{!}{
        \begin{tabular}{lc ccc cc}
            \toprule
            & & \multicolumn{5}{c}{\textsc{$\mathcal{M}^{50}_{TFO}$ (\%)}} \\ 
            \cmidrule(lr){3-7}
            & \textsc{$\mathcal{M}^{0}_{TFO}$ (\%)}& \multicolumn{3}{c}{\textsc{No Uncertainty}} & \multicolumn{2}{c}{\textsc{Uncertainty}} \\ 
            \cmidrule(lr){3-5}
            \cmidrule(lr){6-7}
            & \textsc{No Retrain} & \textsc{No Retrain} & \textsc{Retrained} & \textbf{\textsc{LoCO-PDA}} & \textsc{Retrained} & \textbf{\textsc{LoCO-PDA}} \\
            \midrule
            \textbf{\textsc{City} (185)}           & 74.44  & 65.72 & 68.80 & 69.70 & 60.45 &  68.09 \\
            \textbf{\textsc{Motorway} (26)}        & 72.03  & 63.87 & 71.51 & 72.55 & 57.91 &  69.34 \\
            \textbf{\textsc{Country-side} (204)}   & 74.96  & 66.61 & 69.18 & 70.35 & 63.57 &  69.30 \\
            \textbf{\textsc{Off-road} (26)}        & 78.85  & 70.73 & 85.78 & 86.19 & 77.30 &  82.27 \\
            \bottomrule
        \end{tabular}
    }
    \end{small}
    \caption{On-device domain adaptation accuracy of a 50\% pruned MobileNetV2 network. 
             The table titles are the same as in Table.\ref{tab:acc_comparison}.}
    \label{tab:mobv2_acc_comparison}
    \vspace{-0.5cm}
    \end{minipage}
\end{table}

%% file: tables_tex/further_experiments/uncon_con.tex
\begin{table}[t]
    \begin{minipage}[b]{\linewidth}
    \begin{small}
    \resizebox{\textwidth}{!}{
        \begin{tabular}{l cccc c}
            \toprule
            & \multicolumn{4}{c}{\textbf{\textsc{Top1 Test Accurcay (\%)}}} & \multirow{2}{*}{\textsc{\textbf{Memory (MB)}}}  \\
            \cmidrule(lr){2-5}
            & \textsc{City (185)} & \textsc{Motorway (26)} & \textsc{Country-side (205)} & \textsc{Off-road (26)} & \\
            \midrule
            \textbf{\textsc{Conditional}}   & 73.84 & 76.75 & 74.03 & 87.30 & 10  \\                    
            \textbf{\textsc{Unconditional}} & 73.92 & 74.92 & 74.18 & 88.02 & 400   \\
            \bottomrule
        \end{tabular}
    }
    \end{small}
    \caption{Comparison of retraining accuracy and VAE model memory consumption when using a single CVAE vs 1000 unconditional VAEs.
             The network is an 80\% pruned ResNet50.}
    \label{tab:uncon_con_comparison}
    \vspace{-0.5cm}
    \end{minipage}
\end{table}